\documentclass[letterpaper,10pt,conference]{ieeeconf}

\IEEEoverridecommandlockouts 
\usepackage{times}
\usepackage{multicol}

\usepackage{xparse} 
\usepackage{amsmath} 
\usepackage{amssymb}
\usepackage{amsfonts} 
\usepackage{dsfont}
\usepackage{graphicx}
\usepackage{url}
\usepackage{booktabs}
\usepackage[para]{threeparttable}
\usepackage{multirow}
\usepackage{makecell}
\usepackage{tikz}
\usepackage{soul}

\usepackage{pgfplots}

\usepackage[ruled,vlined]{algorithm2e}
\usepackage{cite} 
\usepackage{hyperref}
\usepackage[capitalize]{cleveref}
\usepackage{xcolor}
\usepackage{framed}
\usepackage{caption}
\usepackage{subcaption}

\SetKwInOut{Parameters}{Parameters}

\usepackage{setspace}

\colorlet{shadecolor}{gray!10}

\NewDocumentCommand\bbm{}{ \begin{bmatrix} }
\NewDocumentCommand\ebm{}{ \end{bmatrix} }
\NewDocumentCommand\Vector{m}{ \boldsymbol{\mathbf{#1}} }
\NewDocumentCommand\Matrix{m}{ \boldsymbol{\mathbf{#1}} }
\NewDocumentCommand\Transpose{m}{ \left.{#1}\right.^{\! T} }

\NewDocumentCommand\Norm{m}{\left\Vert#1\right\Vert }

\NewDocumentCommand\Real{}{ \mathbb{R} }

\NewDocumentCommand\Identity{}{ \Matrix{I} }

\NewDocumentCommand\CoordinateFrame{m}{ \underrightarrow{\Matrix{\mathcal{F}}}_{#1} }

\title{\LARGE \bf
Extrinsic Calibration of 2D Millimetre-Wavelength\\ Radar Pairs Using Ego-Velocity Estimates}

\usetikzlibrary{calc,trees,positioning,arrows,chains,shapes.geometric,%
decorations.pathreplacing,decorations.pathmorphing,shapes,%
matrix,shapes.symbols,plotmarks,decorations.markings,shadows}

\author{Qilong Cheng$^\dagger$, Emmett Wise$^\dagger$, and Jonathan Kelly
\thanks{$\dagger$ denotes equal contribution.}
\thanks{\raggedright All authors are with the Space \& Terrestrial Autonomous Robotics Systems Laboratory at the University of Toronto Institute for Aerospace Studies, Toronto, Canada. Jonathan Kelly is a Vector Institute Faculty Affiliate.
\footnotesize\texttt{<first>.<last>@robotics.utias.utoronto.ca}}
}

\definecolor{red}{RGB}{216,27,96}
\definecolor{blue}{RGB}{30,136,229}
\definecolor{yellow}{RGB}{255,193,7}
\definecolor{green}{RGB}{0,77,64}
\usepackage{tkz-euclide}

\begin{document}
\maketitle

\begin{abstract}
Correct radar data fusion depends on knowledge of the spatial transform between sensor pairs.
Current methods for determining this transform operate by aligning identifiable features in different radar scans, or by relying on measurements from another, more accurate sensor.
Feature-based alignment requires the sensors to have overlapping fields of view or necessitates the construction of an environment map.
Several existing techniques require bespoke retroreflective radar targets.
These requirements limit both where and how calibration can be performed.
In this paper, we take a different approach: instead of attempting to track targets or features, we rely on ego-velocity estimates from each radar to perform calibration.
Our method enables calibration of a subset of the transform parameters, including the yaw and the axis of translation between the radar pair, without the need for a shared field of view or for specialized targets.
In general, the yaw and the axis of translation are the most important parameters for data fusion, the most likely to vary over time, and the most difficult to calibrate manually.
We formulate calibration as a batch optimization problem, show that the radar-radar system is identifiable, and specify the platform excitation requirements.
Through simulation studies and real-world experiments, we establish that our method is more reliable and accurate than state-of-the-art methods.
Finally, we demonstrate that the full rigid body transform can be recovered if relatively coarse information about the platform rotation rate is available.\\
\end{abstract}

\begin{keywords}
	Calibration \& Identification, Radar, Robot Sensing Systems, Sensor Fusion
\end{keywords}

\section{Introduction}

Millimetre-wavelength radar has proven to be a valuable sensing modality for autonomous vehicles (AVs) because radar is relatively robust to inclement weather and is able to provide velocity information \cite{2013_kellner_single}.
However, the field of view of most radar sensors is limited and radars are known to produce noisy range and range-rate measurements.
To provide full coverage of the environment and to ensure redundancy, AVs often aggregate data from multiple radars. Data fusion, in turn, requires accurate knowledge of the spatial transform(s) between the sensors.
The process of determining the sensor-to-sensor transform is known as extrinsic calibration.
While the sensors on board many AVs are factory-calibrated prior to deployment, the spatial transform may change during operation for a variety of reasons (e.g., collisions, wear and tear, etc.).
An ability to perform online, in-situ extrinsic calibration is therefore important for safety and reliability.

Existing extrinsic calibration methods for 2D radar pairs either align clouds of 3D points derived from radar measurements \cite{2021_Olutomilayo_Extrinsic} or involve other (non-radar) sensors that have better accuracy (e.g., lidar units)\cite{2022_burnett_boreas,wise_continuous-time_2021,doer_radar_2020}.
The method in Olutomilayo et al.\ \cite{2021_Olutomilayo_Extrinsic} relies on specialized trihedral radar retroreflectors to ensure sufficient environmental structure for calibration and to simplify the data association problem, for example.
However, these retroreflectors are bespoke and are unavailable outside of the laboratory, limiting the possibility for calibration in the field, during operation.

\begin{figure}
\centering 
\begin{tikzpicture}[
  bluenode/.style={shape=rectangle , draw=white, line width=1},
  yellownode/.style={shape=rectangle , draw=white, line width=1},
  rednode/.style={shape=rectangle, draw=white, line width=1}
]

\node at (-0.428, -0.2) (img) {\includegraphics[scale=0.433]{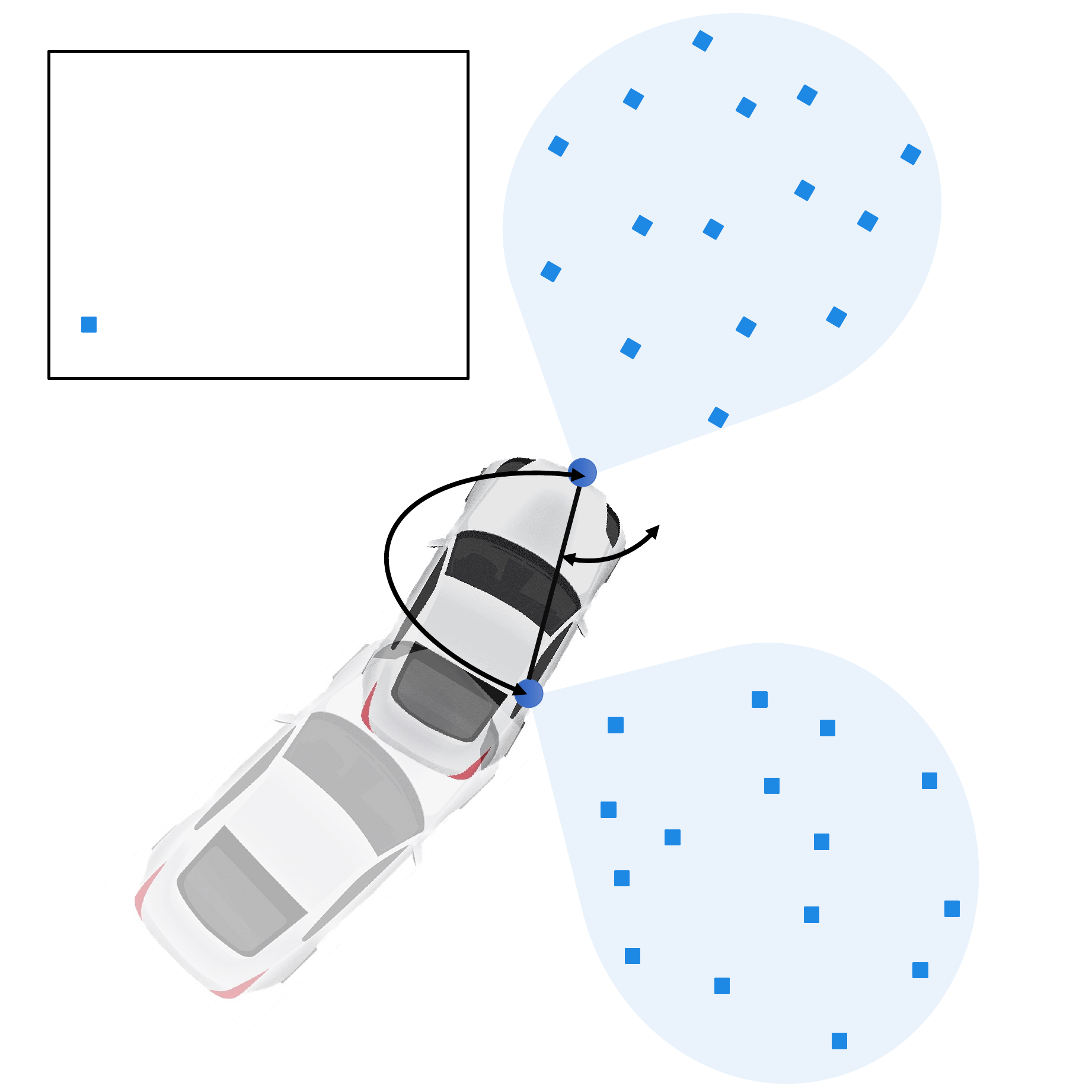}};

\node[anchor=west] (con) [align=left] at (-3.6, 3.05) {Ego-Velocity};
\node[anchor=west] (con) [align=left] at (-3.6, 2.5) {Target Velocity};
\node[anchor=west] (con) [align=left] at (-3.6, 1.95) {Range-Rate};
\node[anchor=west] (con) [align=left] at (-3.6, 1.4) {Targets};

\draw[red][ ultra thick] (-3.920,3.075) -- (-3.72,3.075);
\draw[blue][ ultra thick] (-3.920,2.525) -- (-3.72,2.525);
\draw[yellow!80!orange][ ultra thick] (-3.920,1.975) -- (-3.72,1.975);

\draw [-][ultra thick] (-0.15, 0.35) -- (-0.58, -1.32);

\draw [-stealth][thick] (-0.15, 0.35) -- (1.9, 3.814)  node[below right] {y};
\draw [-stealth][thick] (-0.986, 0.83) -- (0.716,-0.24)  node[right] {x};

\draw [red][-stealth][ultra thick] (-0.15, 0.35) -- (0.36, 1.217);

\draw[dotted][ultra thick] (-0.15, 0.35) -- (2.72, 2.88);
\draw[dotted][ultra thick] (-0.15, 0.35) -- (2.675, 2.05);
\draw[dotted][ultra thick] (-0.15, 0.35) -- (0.305, 3.7);
\draw[dotted][ultra thick] (-0.15, 0.35) -- (-0.55, 2.8);

\draw[blue][-stealth][ultra thick] (1.5, 3.13) -- (1, 2.264);
\draw[yellow!80!orange][-stealth][ultra thick] (1.5, 3.13) -- (1, 2.264);

\draw[blue][-stealth][ultra thick] (1.95, 2.226) -- (1.45, 1.36);
\draw[yellow!80!orange][-stealth][ultra thick] (1.95, 2.20) -- (1.26, 1.594);
\draw[dotted][ultra thick] (1.24, 1.61) -- (1.45, 1.36);

\draw[blue][-stealth][ultra thick] (-0.38, 1.85) -- (-0.88, 0.984);
\draw[yellow!80!orange][-stealth][ultra thick] (-0.39, 1.82) -- (-0.27, 1.08);
\draw[dotted][ultra thick] (-0.88, 0.984) -- (-0.263, 1.09);

\draw[blue][-stealth][ultra thick] (0.25, 3.11) -- (-0.33, 2.244);
\draw[yellow!80!orange][-stealth][ultra thick] (0.225, 3.09) -- (0.097, 2.18);
\draw[dotted][ultra thick] (-0.35, 2.234) -- (0.07, 2.185);

\draw[blue][-stealth][ultra thick] (1.76, 1.52) -- (1.26, 0.654);
\draw[yellow!80!orange][-stealth][ultra thick] (1.76, 1.50) -- (1.03, 1.06);
\draw[dotted][ultra thick] (1.26, 0.654) -- (1.01, 1.06);

\draw [red][-stealth][ultra thick]  (-0.56, -1.32) -- (-0.06, -0.454);

\draw [-stealth][thick]  (-0.58, -1.32) -- (2.874,-3.32)  node[below] {y};
\draw [-stealth][thick] (-0.16, -0.627) -- (-1.17, -2.359)  node[anchor=north west] {x};

\draw[dotted][ultra thick]  (-0.58, -1.32) -- (2.62, -1.378);
\draw[blue][-stealth][ultra thick] (1.17, -1.33) -- (0.67, -2.196); 
\draw[dotted][ultra thick] (0.68, -2.216) -- (0.70, -1.3461);
\draw[yellow!80!orange][-stealth][ultra thick] (1.17, -1.352) -- (0.68, -1.3411);

\draw[dotted][ultra thick]  (-0.58, -1.32) -- (2.62, -1.98);
\draw[blue][-stealth][ultra thick] (2.42, -1.92) -- (1.92, -2.786); 
\draw[dotted][ultra thick] (1.92, -2.786) -- (2.12, -1.84);
\draw[yellow!80!orange][-stealth][ultra thick] (2.43, -1.94) -- (2.12, -1.875);

\draw[dotted][ultra thick]  (-0.58, -1.32) -- (2.62, -3.56);
\draw[blue][-stealth][ultra thick] (2.365, -3.355) -- (1.865, -4.141); 
\draw[dotted][ultra thick] (1.865, -4.141) -- (2.41, -3.38);
\draw[yellow!80!orange][-stealth][ultra thick] (2.385, -3.375) -- (2.42, -3.40);

\draw[dotted][ultra thick]  (-0.58, -1.32) -- (1.32, -4.12);
\draw[blue][-stealth][ultra thick] (0.89, -3.486) -- (0.39, -4.352); 
\draw[dotted][ultra thick] (0.39, -4.352) -- (1.145, -3.86);
\draw[yellow!80!orange][-stealth][ultra thick] (0.89, -3.486) -- (1.145, -3.86);

\node (con) [align=center] at (2.42, 0.35) {Radar $a$ \\ Point Cloud \\ ($r_{a,i}$, $\theta_{a,i}$, $\dot{r}_{a,i})$};
\node (con) [align=center] at (-0.75, -3.75) {Radar $b$ \\ Point Cloud \\ ($r_{b,i}$, $\theta_{b,i}$, $\dot{r}_{b,i})$};
\node (con) [align=center] at (-3, 0) {Rigid Body};
\draw [black][-stealth][ultra thick] (-3, -0.25) -- (-2.5, -1.5);

\node (con) [align=center] at (-1.6, 0.4) {$\Matrix{R}$};
\node (con) [align=center] at (0.3, -0.45) {$\theta_t$};

\node (con) [align=center] at (-1.76, -0.75) {$t_j$};
\node (con) [align=center] at (-3.2, -1.95) {$t_{j-1}$};

\end{tikzpicture}
\vspace{1mm}
\caption{Illustration of our 2D radar-to-radar extrinsic calibration problem. Radars $a$ and $b$ measure the range, azimuth, and range-rate to targets in the environment. Assuming that the targets are stationary in a fixed, world reference frame, we can estimate the ego-velocity of each radar. Our algorithm fuses the ego-velocity estimates from radars $a$ and $b$ to estimate the rotation, $\Matrix{R}$, and translation direction $\theta_t$ between the the two sensors.}
\label{fig:main}
\vspace{-4mm}
\end{figure}
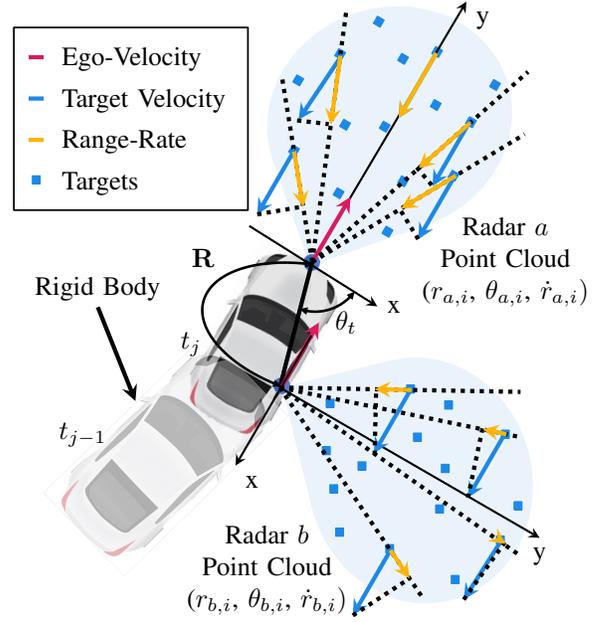

In this paper, we study the radar-to-radar extrinsic calibration problem and develop a calibration approach that does not require specialized targets or other sensors.
To avoid the challenges of data association, we instead propose a method that relies on instantaneous ego-velocity estimates from each radar only, as shown in \Cref{fig:main}.
More specifically, we determine the rotation angle and translation axis (i.e., unit vector) between the sensors. Radar data fusion, especially radar ego-velocity fusion, is most sensitive to these parameters. 
Under nominal operating conditions, the distance between the sensors is unlikely to change appreciably from specifications.
In contrast, the orientation of each radar can easily be altered (e.g., by a minor impact) and is very difficult to measure by hand.
To the best of the authors' knowledge, this is the first work to explore extrinsic calibration of sensor pairs that provide velocity estimates only.
That is, no direct information about the rotation or rotation rate of the sensor platform is considered (instead, rotation must be inferred from the velocities).
Herein, we:
\begin{enumerate}
    \item show that the yaw angle and the direction of translation between pairs of coplanar 2D radar units can be determined from ego-velocity estimates only; 
    \item formulate a batch solver for the calibration problem and prove that the parameters are identifiable, given sufficient excitation;
    \item carry out simulation studies to analyze the sensitivity of our method to varying levels of measurement noise;
    \item confirm via real-world experiments that our approach is more accurate and reliable than two state-of-the-art methods; and
    \item demonstrate that the full spatial transform can be recovered when an additional, coarse source of information about the platform rotation rate is available.
\end{enumerate}

The remainder of the paper is organized as follows.
In \Cref{sec:related-work}, we review existing radar extrinsic calibration algorithms.
\Cref{sec:problem} formulates our batch optimization problem.
We prove that the calibration parameters are identifiable and establish the necessary trajectory excitation requirements in \Cref{sec:observability}.
In \Cref{sec:exp}, we show, through simulations and real-world experiments, that our algorithm is more reliable and accurate at estimating the yaw angle and the translation axis between radar pairs than two state-of-the-art-methods.
Finally, we summarize our work and discuss future research directions in \Cref{sec:conclusion}. 

\section{Related Work}
\label{sec:related-work}

In this section, we survey pairwise extrinsic calibration methods where one (or both) of the sensors in the pair is a mm-wavelength radar.
Sections \ref{subsec:target} and \ref{subsec:targetless} detail target-based and target-free extrinsic calibration algorithms, respectively, that rely on feature detection and matching.
In \Cref{subsec:ego-cal}, we review extrinsic calibration methods that relate the instantaneous radar ego-velocity to the motion of the second sensor.

\subsection{Target-Based Methods}
\label{subsec:target}

Most existing radar-camera extrinsic calibration algorithms estimate the projective transform (homography) between the horizontal 2D radar sensing plane and the camera image plane.
Due to the sparse and noisy nature of radar measurements, these methods rely on specialized trihedral retroreflectors that produce point-like detections in the radar and camera data while simplifying the correspondence problem \cite{sugimoto_obstacle_2004,Wang2011,kim_data_2014,kim_radar_2018}.
Additionally, although 2D radars are incapable of estimating elevation, the sensors do often detect off-plane targets, and these detections bias the homography estimate.
Since signal returns from targets on the radar horizontal plane are stronger than those from off-plane targets, Sugimoto et al.\ \cite{sugimoto_obstacle_2004} filter on-plane targets by maximizing the measured radar cross section (RCS) of the targets.

The most common error metric for radar-to-sensor extrinsic calibration is a form of  `reprojection' error, which defines the misalignment between identifiable targets (objects) viewed by both sensors.
For example, Olutomilayo et al.\ \cite{2021_Olutomilayo_Extrinsic} estimate the 2D transform between the radar frame and a vehicle coordinate frame by aligning radar measurements of stationary retroreflectors with a known map in the vehicle coordinate frame.
The approaches of El Natour et al. \cite{elnatour_radar_2015}, Domhoff et al.\ \cite{Domhof2019_calibration}, and Per\u si\'c et al.\ \cite{Persic2019_calibration} treat all radar measurements as lying on spherical arcs with constant range and azimuth (i.e., the measurements vary only in elevation).
To estimate the 3D transform between sensor pairs, the arcs are aligned with the measurements from a second sensor. 
In order to account for the elevations of the retroreflector targets relative to the horizontal radar sensing plane, these methods introduce additional calibration constraints by designing specific target arrangements \cite{elnatour_radar_2015, Domhof2019_calibration} or explicitly modelling the radar-target interactions \cite{Persic2019_calibration}. 
All of these techniques require the sensor pair to simultaneously view one or more specialized targets, so the sensors must share overlapping fields of view.
Our approach does not require specialized infrastructure or a shared field of view, allowing for calibration of a wider range of sensor configurations and in more environments.

\subsection{Target-Free Methods}
\label{subsec:targetless}

In contrast to methods that rely on specialized radar retroreflectors, target-free or `targetless' algorithms estimate the radar-to-sensor transform by aligning identifiable environment features (observed by both sensors). 
Sch\"oller et al. \cite{scholler} train a neural network to correct an inaccurate rotation estimate between a 2D radar-camera pair using raw camera and radar vehicle detection data.
Per\u si\'c et al. \cite{persic_online_2021} align tracked objects to determine the yaw angle between 2D radar-camera and radar-lidar pairs.
Due to the challenge of tracking environmental features consistently across radar scans, these methods only calibrate the rotation between the sensors.
Burnett et al.\ \cite{2022_burnett_boreas} estimate the transform between a 2D radar-lidar pair, where both sensors have a 360$^\circ$ field of view.
The method in \cite{2022_burnett_boreas} aligns measured radar and lidar point clouds, which requires a large number of jointly-observed features.
Heng et al.\ \cite{heng_automatic_2020} estimate the extrinsic calibration parameters between 3D radar-lidar pairs by constructing a lidar point cloud map and localizing the 3D radar units within the map.
While this approach could possibly be used for 2D radars, the method requires the known poses of the vehicle and the construction of dense map.
Our approach does not require tracking of environmental features, which  simplifies the calibration process.

\subsection{Ego-Velocity Methods}
\label{subsec:ego-cal}

Ego-velocity methods estimate the transform by minimizing the error between radar ego-velocity estimates and the motion of another sensor  \cite{2013_kellner_single,Stahoviak_Velocity_2019}.
By minimizing the lateral velocity error between a radar and an inertial measurement unit (IMU), Kellner et al.\ \cite{Kellner2015_calibration} estimate the rotation angle between a radar-IMU pair, but their scheme requires accurate knowledge of the translation between the sensors.
Doer et al.\ \cite{doer_radar_2020} and Wise et al.\ \cite{wise_continuous-time_2021} extend this approach to 3D radar-IMU and radar-camera extrinsic calibration. 
To date, each ego-velocity method relies on one sensor that provides rotation information (e.g., angular velocity or $\mathrm{SO}(2)$ measurements relative to an inertial frame).
Herein, we determine the subset of extrinsic calibration parameters that are identifiable without angular velocity measurements.

\section{Problem Formulation}
\label{sec:problem}

\subsection{Notation} 

In this paper, Latin and Greek letters, such as $a$ and $\alpha$, represent scalar variables.
Lowercase (e.g., \textbf{$\Vector{h}$} and \textbf{$\Vector{\sigma}$}) and uppercase (e.g., $\Matrix{\Theta}$ and $\Matrix{C}$) boldface characters are reserved for vectors and matrices, respectively.
A Cartesian reference frame is identified by $\CoordinateFrame{}$. 
The translation vector from $\CoordinateFrame{a}$ to $\CoordinateFrame{b}$, expressed in $\CoordinateFrame{a}$, is denoted by $\Vector{t}_a^{ba}$.
The function $\Matrix{R}(\theta)$ maps $\theta \in \Real$ to an element of $\mathrm{SO}(2)$; for example, $\Matrix{R}(\theta_{ab})$ defines the rotation from $\CoordinateFrame{b}$ to $\CoordinateFrame{a}$.
We use $\Identity_{n}$ to denote the $n$-by-$n$ identity matrix. 
The operator $\times$ is the cross product operator.
The unary operator $^\wedge$ acts on $r \in \Real$ to produce 
\begin{equation}
r^\wedge = \bbm 0 & -r \\ r & 0 \ebm.
\end{equation} 

\subsection{Radar Ego-Velocity Estimation}

Let the static world and moving radar frames be $\CoordinateFrame{w}$ and $\CoordinateFrame{r}$, respectively.
At each time index $j$, the radar detects $N$ stationary, environmental features in $\CoordinateFrame{w}$.
The resulting radar measurement is $\{\bbm r_1 & \theta_1& \dot{r}_1 \ebm, \dots, \bbm r_N & \theta_N& \dot{r}_N \ebm\}$, where $r_i$, $\theta_i$, and $\dot{r}_i$ are, respectively, the range, azimuth, and range-rate (i.e., the Doppler velocity) of feature $i$ in $\CoordinateFrame{r}$.
If we assume that each feature lies on the horizontal plane of the radar, then the range-rate of a feature is
\begin{equation}
\dot{r}_i = - \bbm \sin(\theta_i) & \cos(\theta_i)\ebm\Vector{h}_r^j,
\end{equation}
where $\Vector{h}_r^j$ is the 2D radar ego-velocity at time $j$.
A depiction of the relationship between the range-rate of stationary features and ego-velocity of the radar is shown in \Cref{fig:main}.  

Radar ego-velocity estimation can be cast as a linear least squares problem, where the measurement model is
\begin{equation}
\underbrace{
\begin{bmatrix}
-\dot{r}_1 \\
-\dot{r}_2 \\ 
\vdots \\
-\dot{r}_N
\end{bmatrix} = 
\begin{bmatrix}
\sin(\theta_1) & \cos(\theta_1)  \\
\sin(\theta_2) & \cos(\theta_2)  \\
\vdots  \\
\sin(\theta_N) & \cos(\theta_N) 
\end{bmatrix}
\Vector{h}_r^j
}_{\Vector{y} = \Matrix{A}\Vector{h}_r^j}.
\end{equation}
The resulting error equation is 
\begin{equation}
\Vector{\epsilon}  = \Vector{y} - \Matrix{A}\Vector{h}_r^j,
\end{equation}
and the radar ego-velocity estimation problem is
\begin{equation}
\min_{\Vector{h}_r^j \in \Real^2} \Transpose{\Vector{\epsilon}}\Vector{\epsilon}. 
\end{equation} 
As a result, the estimated ego-velocity at time $j$ is
\begin{equation} \label{eqn:ego-soln}
{\Vector{h}_r^j}^\star = (\Transpose{\Matrix{A}}\Matrix{A})^{-1}\Transpose{\Matrix{A}} \Vector{y},
\end{equation}
with covariance
\begin{equation} \label{eqn:ego-vel-cov}
\Matrix{\Sigma}_r^j = \frac{(\Transpose{\Vector{\epsilon}}\Vector{\epsilon})(\Transpose{\Matrix{A}}\Matrix{A})}{N - 2}.
\end{equation}
We leverage RANSAC to remove outliers such as targets moving relative to $\CoordinateFrame{w}$ and multipath radar reflections \cite{richards_principle_2010}. 

\subsection{Radar Ego-Velocity Measurement Models} 

Let $\CoordinateFrame{a}$ and $\CoordinateFrame{b}$ be the reference frames of two rigidly attached radars that share and move along one horizontal sensing plane. 
The ego-velocity measurement models for radars $a$ and $b$, at time $j$, are 
\begin{align}
\Vector{h}_{r,a}^j = & \Vector{v}_{r,a}^j + \Vector{n}_{r,a}^j,\label{eqn:radarmdl-a}\\
\Vector{n}_{r,a}^j \sim & \mathcal{N}\left(\Vector{0},\Matrix{\Sigma}_{r,a}^j\right), \nonumber\\
\Vector{h}_{r,b}^j = & \Matrix{R}(\theta_{ba})({\omega^j}^\wedge \Vector{t}_a^{ba} + \Vector{v}_{r,a}^j) + \Vector{n}_{r,b}^j,\label{eqn:radarmdl-b}\\
\Vector{n}_{r,b}^j \sim & \mathcal{N}\left(\Vector{0},\Matrix{\Sigma}_{r,b}^j\right), \nonumber
\end{align}
where $\Vector{v}_{r,a}$ is the ego-velocity of the radar, $\theta_{ba}$ is the rotation from radar $a$ to radar $b$, $\omega^j$ is the angular velocity of the rigid body, and $\Vector{t}_a^{ba}$ is the translation from radar $a$ to radar $b$, expressed in the reference frame of $a$.
The vectors $\Vector{n}_{r,a}^j$ and $\Vector{n}_{r,b}^j$ are additive zero-mean Gaussian noise terms with covariances $\Matrix{\Sigma}_{r,a}^j$ and $\Matrix{\Sigma}_{r,b}^j$, respectively.
The values of $\Matrix{\Sigma}_{r,a}^j$ and $\Matrix{\Sigma}_{r,b}^j$ are determined with use of \Cref{eqn:ego-vel-cov}.

The error equations corresponding to the ego-velocity estimates are
\begin{equation} \label{eqn:errors}
\begin{aligned}
\Vector{e}_{r,a}^j = & \Vector{h}_{r,a}^j - \Vector{v}_{r,a}^j,\\
\Vector{e}_{r,a}^j \sim & \mathcal{N}\left(\Vector{0},\Matrix{\Sigma}_{r,a}^j\right),\\
\Vector{e}_{r,b}^j = & \Vector{h}_{r,b}^j - \Matrix{R}(\theta_{ba})({\omega^j}^\wedge \Vector{t}_a^{ba} + \Vector{v}_{r,a}^j),\\
\Vector{e}_{r,b}^j \sim & \mathcal{N}\left(\Vector{0},\Matrix{\Sigma}_{r,b}^j\right),
\end{aligned}
\end{equation}
where $\Vector{h}_{r,a}^j$ and $\Vector{h}_{r,b}^j$ are the values from \Cref{eqn:ego-soln} for radars $a$ and $b$, respectively.

\subsection{Batch 2D Radar to Radar Extrinsic Calibration}

Given $M$ pairs of synchronized radar measurements, the vector of parameters that we wish to estimate includes the ego-velocity of radar $a$ from time $1$ to $M$, the angular velocity of radar $a$ from time $1$ to $M$, the translation from radar $a$ to $b$ expressed in $\CoordinateFrame{a}$, and the rotation from radar $\CoordinateFrame{a}$ to $\CoordinateFrame{b}$,
\begin{equation}
\begin{aligned}
{\Vector{x}}^{\! T} = & \left[\begin{matrix}
{\Vector{v}_{r,a}^1}^{\! T} & \! \omega^1 & \! \cdots & \! \Transpose{\Vector{v}_{r,a}^{M}} & \! \omega^{M}
\;\;\; {\Vector{t}_a^{ba}}^{\! T} & \! \theta_{ba}
\end{matrix}\right].
\end{aligned}
\end{equation}
Our calibration problem is to solve
\begin{equation}\label{prob:opt-base}
\min_{\Vector{x}} \sum_{j=1}^{M} \Transpose{\Vector{e}_{r,a}^j}{\Matrix{\Sigma}_{r,a}^j}^{-1}\Vector{e}_{r,a}^j + \Transpose{\Vector{e}_{r,b}^j}{\Matrix{\Sigma}_{r,b}^j}^{-1}\Vector{e}_{r,b}^j.
\end{equation}

\subsection{Scale and Angular Velocity Indistinguishability}
\label{subsec:scale}

Unfortunately, the optimization problem defined by \Cref{prob:opt-base} has infinitely many indistinguishable solutions.
Given any solution that minimizes \Cref{prob:opt-base}, another minimizer can be found by arbitrarily scaling $\omega^j \; \forall \; j=1,\dots,M$ and $\Vector{t}_a^{ba}$ by $\gamma \in \Real$ and $\frac{1}{\gamma}$, respectively.
However, the problem can be made distinguishable with additional constraints.

To make the optimization problem identifiable (see \Cref{sec:observability}), we constrain
\begin{equation}
\Norm{\Vector{t}_a^{ba}}_2 = 1.
\end{equation}
We enforce this constraint by setting
\begin{equation} \label{eqn:constraints}
\Vector{t}_a^{ba} = \bbm \cos(\theta_t) \\ \sin(\theta_t) \ebm,
\end{equation}   
where $\theta_t$ is the angle from the x-axis of radar $a$ to the line of possible translations between radars $a$ and $b$.
Since the angle to the line is periodic with period $\pi$, we bound $0 \leq \theta_t < \pi$. 
We denote the resulting unscaled angular velocity as $\omega_\gamma^j$.
Consequently, our vector of parameters for the optimization problem becomes 
\begin{equation} \label{eqn:new_state}
\Transpose{\Vector{x}} = \left[\begin{matrix}
{\Vector{v}_{r,a}^1}^{\! T} & {\omega_\gamma^1} & \cdots & {\Vector{v}_{r,a}^M}^{\! T} & \omega_\gamma^M & \theta_t & \theta_{ba} \end{matrix}\right].
\end{equation} 
We substitute \Cref{eqn:constraints} into \Cref{prob:opt-base} and solve the problem using the Levenberg-Marquardt algorithm.

\subsection{Problem Initialization}

Since the two radar sensors provide no rotational information, we require a method to initialize the angular velocities that appear in \Cref{eqn:new_state}.
To start, we determine $\theta_{ba}$ by finding the $K$ pairs of ego-velocity estimates that have similar magnitudes.
Using these `velocity pairs,' we compute
\begin{equation}
\theta_{ba}^l =\bbm 0 & 0 & 1 \ebm \left(\bbm \Vector{h}_{r,a}^l \\ 0 \ebm \times \bbm \Vector{h}_{r,b}^l \\ 0 \ebm\right).
\end{equation}
The initial $\theta_{ba}$ is the median of $\theta_{ba}^l \; \forall \; l=1,...,K$.
To initialize $\theta_t$, we use
\begin{equation}
\begin{aligned}
\Vector{b}^j =& \bbm b^j_x & b^j_y \ebm^{\! T} = \Transpose{\Matrix{R}(\theta_{ba})}\Vector{h}_{r,b}^j - \Vector{h}_{r,a}^j, \\
\theta_t^j = & \text{arctan2}\left(\frac{b^j_y}{\Norm{\Vector{b}^j}_2}, -\frac{b^j_x}{\Norm{\Vector{b}^j}_2}\right).
\end{aligned}
\end{equation}
Each $\theta_t^j$ is mapped to the corresponding value within $\left[0,\pi\right)$ and the initial $\theta_t$ is the median of $\theta_t^j \; \forall \; j=1,...,M$.
By fixing $\theta_t$ and $\theta_{ba}$ to our initial estimates, \Cref{prob:opt-base} becomes an unconstrained quadratic problem. 
We solve this problem to initialize $\omega_\gamma^j \; \text{and} \; \Vector{v}_{r,a}^j \; \forall \; j=1\dots M$.

\section{Identifiability}
\label{sec:observability}

In this section, we prove that, given sufficient excitation of the system, the extrinsic calibration problem is identifiable.
Since a problem that is locally weakly observable is also identifiable (in the batch setting), we use the rank criterion defined by Hermann and Krener in \cite{hermann_nonlinear_1977} in our proof.
\Cref{sec:obs-crit} reviews the rank criterion.
In \Cref{sec:obs-proof}, we demonstrate that our problem is locally weakly observable.
Finally, we highlight important degenerate motions in \Cref{subsec:obs-degen} that result in a loss of observability and potentially also identifiability.

\subsection{The Observability Rank Criterion}
\label{sec:obs-crit}

Consider the system
\begin{equation}
S\,\begin{cases}
\dot{\Vector{x}} = \Vector{f}_0(\Vector{x}) + \sum_{j=1}^p \Vector{f}_j(\Vector{x})\,u_j \\
\Vector{y} = \Vector{h}(\Vector{x})
\end{cases},
\end{equation}
where $\Vector{x}$ is the state vector, $\Vector{f}_0(\Vector{x})$ is the drift vector field, $\Vector{f}_j(\Vector{x})$ is a vector field on the state manifold that is linear with respect to the control input $u_j$, $\Vector{y}$ is the measurement vector, and $\Vector{h}(\Vector{x})$ is the measurement model.
Given the vector field $\Vector{f}(\Vector{x})$, we can compute the Lie derivative of $\Vector{h}$ with respect to $\Vector{f}$, which is defined as 
\begin{equation}
L_{\Vector{f}} \Vector{h}(\Vector{x}) = \nabla_{\Vector{f}}\Vector{h}(\Vector{x}) = \frac{\partial \Vector{h}(\Vector{x})}{\partial \Vector{x}}\Vector{f}(\Vector{x}).
\end{equation}
The $n^\text{th}$ Lie derivative of $h$ with respect to $\Vector{x}$ along vector field $\Vector{f}$ is defined as,
\begin{equation}
L_{\Vector{f}}^n \Vector{h}(\Vector{x}) = \frac{\partial L_{\Vector{f}}^{n-1} \Vector{h}(\Vector{x})}{\partial \Vector{x}}\Vector{f}(\Vector{x}),
\end{equation}  
where $L^0 \Vector{h}(\Vector{x})=\Vector{h}(\Vector{x})$.

The Lie derivatives can be vertically stacked to form the observability matrix $\Matrix{O}$.
From Hermann and Krener \cite{hermann_nonlinear_1977}, a system is locally weakly observable at $\Vector{x}$ if the matrix $\Matrix{O}$ is full column rank at $\Vector{x}$.

\subsection{Observability of Extrinsic Calibration}
\label{sec:obs-proof}

Let $\Vector{h}_{r,a}$ and $\Vector{h}_{r,b}$ be $2$D radar ego-velocity measurements. 
We define the state at timestep $t_j$ as
\begin{equation}
\Vector{x}^T(t_j) = \bbm \omega_\gamma(t_j) & \alpha_\gamma(t_j) & \theta_t & \theta_{ba}\ebm,
\end{equation}
where $\alpha_\gamma(t_j)$ is the instantaneous angular acceleration.\footnote{The analysis can, in fact, be simplified by removing the angular acceleration state; we use this formulation, specifically, in \Cref{subsec:obs-degen}.}
We assume the vehicle follows the constant angular acceleration model given by
\begin{equation}
\begin{aligned}
\dot{\alpha}_\gamma(t_j) =& n_\alpha, \\
n_\alpha \sim & \mathcal{N}\left(0, \sigma_\alpha^2 \right),
\end{aligned}
\end{equation} 
where $n_\alpha$ is an additive zero-mean Gaussian noise term with variance $\sigma_\alpha^2$.
Since the motion is noiseless in this analysis (i.e., $\dot{\alpha}_\gamma(t_j) = 0$), the motion model is 
\begin{equation}
\dot{\Vector{x}}^T(t_j) = \bbm  \alpha_\gamma(t_j) & \Vector{0}_{3\times 1}^T\ebm.
\end{equation}
We can substitute \Cref{eqn:radarmdl-a} into \Cref{eqn:radarmdl-b}, which simplifies the measurement model to
\begin{equation}
\Vector{h}_{r,b}^j = \Matrix{R}(\theta_{ba})\left(\omega_\gamma(t_j)^\wedge\bbm \cos\theta_t \\ \sin\theta_t \ebm + \Vector{h}_{r,a}^j\right).
\end{equation}
The observability matrix of this system can be written as
\begin{equation}
\Matrix{O}=\bbm \nabla_{\Vector{x}} L^0\Vector{h}_{r,b}^j \\ \nabla_{\Vector{x}} L^1_f\Vector{h}_{r,b}^j \ebm,
\end{equation}
which is full column rank\footnote{The rank of $\Matrix{O}$ can be determined using a symbolic math package. We omit the full proof for brevity.} except when the sensor platform motion is degenerate, as discussed below.

\subsection{Degeneracy Analysis}
\label{subsec:obs-degen}

The system is unobservable (and potentially unidentifiable) when $\Matrix{O}$ does not have full column rank.
The determinant of the observability matrix is
\begin{equation}
\det(\Matrix{O}) = \bbm 0 & 0 & \alpha_\gamma(t_j) \ebm \left(\bbm \Vector{h}_{r,a}^j \\[1mm] 0 \ebm \times \bbm \cos\theta_t \\ \sin\theta_t \\ 0 \ebm\right),
\end{equation}
which is rank-deficient when $\det(\Matrix{O})=0$.
As a result, the system must have nonzero angular acceleration, $\alpha_\gamma$, and nonzero ego-velocity, $\Vector{h}_a$.
Additionally, the direction of ego-motion must not align with the sensor translation axis.

\section{Experiments}
\label{sec:exp}

To verify the performance of our algorithm, we conducted a series of simulated and real-world experiments.
In \Cref{subsec:sim} we show, using simulated data, that our algorithm is robust to realistic levels of radar measurement noise and that it yields an improved ego-velocity estimate.
In \Cref{subsec:real}, we compare our approach to two state-of-the-art methods on the publicly-available Endeavour dataset.\footnote{The dataset is available at: \url{ https://gloryhry.github.io/2021/06/25/Endeavour_Radar_Dataset.html}}

\subsection{Simulation Studies}
\label{subsec:sim}

\begin{figure}[t!]
\centering 
\includegraphics[width=\columnwidth]{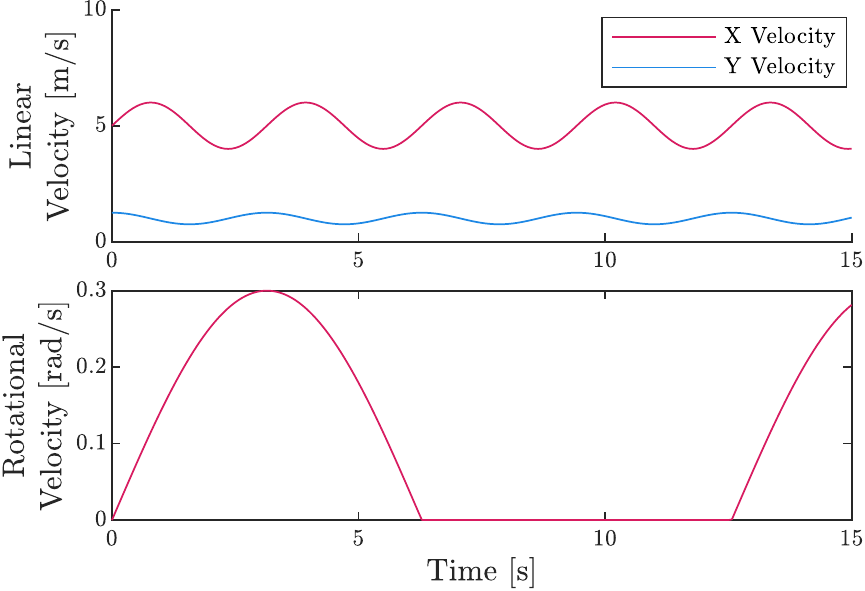}
\caption{Top: ego-velocity of radar $a$ over 15 s. Bottom: angular velocity of radar $a$ over 15 s. Both plots show the full period of the  velocity functions.}
\label{fig:sim_traj}
\vspace{-5mm}
\end{figure}

We performed a series of simulation studies to evaluate the robustness of our algorithm to measurement noise. 
We varied the simulation duration and the level of noise and generated 100 randomized trials with each pair of settings.
Each simulation ranged in duration from 15 s to 120 s; the simulated sensor platform followed a periodic, nominal (noise-free) trajectory with sufficient excitation for our calibration problem (see \Cref{fig:sim_traj}).
The radar ego-velocity estimates for radars $a$ and $b$ were computed using the ground truth linear and angular velocities of the platform along the trajectory.
Ego-velocity measurements from radars $a$ and $b$ were then corrupted with zero-mean Gaussian noise ($\Matrix{\Sigma}_{r,a}^j=\Matrix{\Sigma}_{r,b}^j=\sigma_r^2\Identity_2$), where the standard deviation of the noise ($\sigma_r$) ranged from 0.05 m/s to 0.2 m/s. 
Based on our experiments (discussed in more detail in \Cref{subsec:real}), we found the real-world measurement noise to be at the lower end of this range.

\begin{figure}[b!]
\vspace{-5mm}
\centering 
\includegraphics[width=\columnwidth]{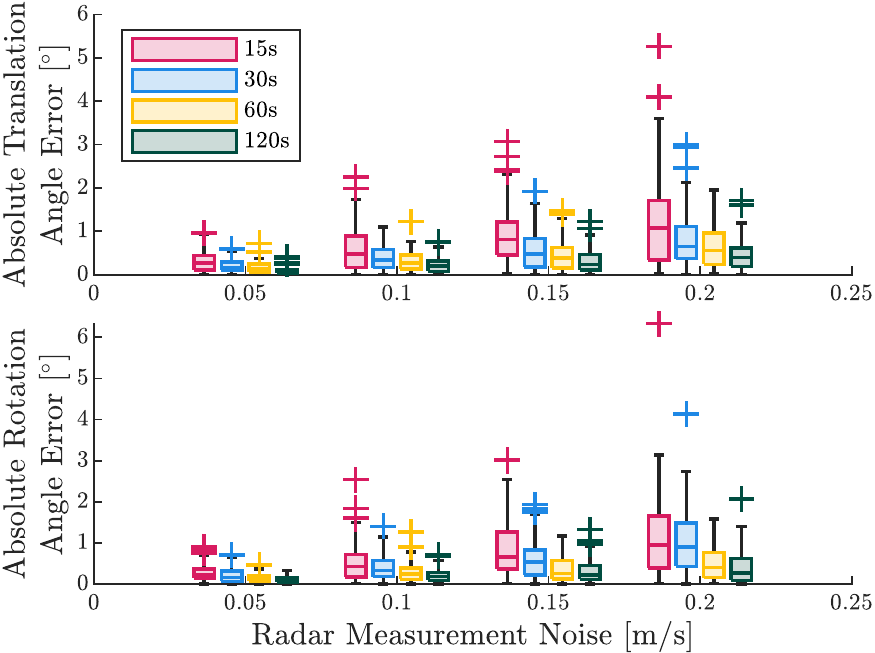}
\caption{Absolute translation direction and rotation angle error for our algorithm at varying levels of measurement noise and simulation durations. The translation direction is the angle from the x-axis of radar $a$ to the line of indistinguishable translations between radars $a$ and $b$.}
\label{fig:sim_res}
\end{figure}

\begin{figure}[t!]
\centering 
\includegraphics[width=\columnwidth]{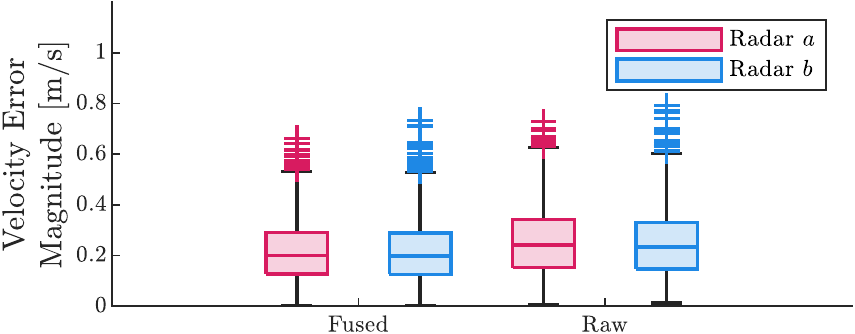}
\caption{Raw and fused ego-velocity estimate errors of radars $a$ and $b$ for an experiment that is 120 s in duration with a measurement noise level of 0.2 m/s.}
\label{fig:vel_res}
\vspace{-2mm}
\end{figure}

The error distribution of the estimated calibration parameters is shown in \Cref{fig:sim_res}.
For most noise levels and durations, our estimated translation direction and rotation angle are, respectively, within 2$^\circ$ and 3$^\circ$ of the ground truth.
\Cref{fig:vel_res} shows that the median of the estimated ego-velocity errors for radars $a$ and $b$ are both 4 cm/s lower than the raw estimates.
Importantly, this improvement can be achieved without the need for additional rotation information.

\subsection{Real-World Experiments}
\label{subsec:real}

We demonstrate the reliability of our method and compare to two state-of-the-art algorithms on the Endeavour dataset.
Post-hoc extrinsic calibration for this dataset is challenging because the environments contain no trihedral reflectors.
We demonstrate that the lack of trihedral reflectors has a negligible impact on our method, but is detrimental to the method in Olutomilayo et al.\ \cite{2021_Olutomilayo_Extrinsic}.
Additionally, we show that the parameters estimated by our method result in smaller velocity errors than the parameters estimated by two state-of-the-art methods.
The first technique follows the approach in Olutomilayo et al.\ \cite{2021_Olutomilayo_Extrinsic}.
To build the required map for this method, we collate measurements from one radar while the vehicle is stationary.
The second method is similar to the approach in Burnett et al.\ \cite{2022_burnett_boreas}.
The parameters estimated by this method are included in the Endeavour dataset.\footnote{The radar to lidar extrinsic calibration code for the Endeavour dataset can be found at: \url{https://github.com/gloryhry/radar_lidar_static_calibration}}
Next, we demonstrate that normal driving motions provide sufficient excitation to calibrate radar pairs that have translation axes which align with the forward direction of the vehicle.
Finally, we show that, when a source of angular velocity information is available, the scale of the translation between the radar pair can be estimated.

The Endeavour dataset was collected from a small shuttle bus driving around three different loops in a campus setting.
The dataset contains two runs for each loop, where each run is roughly 10 minutes long.
The shuttle bus has a BDStar Navigation Npos320 RTK GNSS, four Velodyne VLP-16 lidars, and five Continental ARS430 radars, which operate at 100 Hz, 10 Hz, and 14 Hz, respectively.
The radar labelled \texttt{Near\_5} is mounted on the front bumper and observes the environment in front of the vehicle.
Radar pairs \texttt{Near\_3}--\texttt{Near\_1} and \texttt{Near\_4}--\texttt{Near\_2} observe the environment surrounding the sides of the vehicle.
Radars \texttt{Near\_3} and \texttt{Near\_1} are mounted on the front and back driver's side of the vehicle, while radars \texttt{Near\_4} and \texttt{Near\_2} are mounted on the front and back passenger side of the vehicle.\footnote{Additional detailed information is available at: \url{ https://gloryhry.github.io/2021/06/25/Endeavour_Radar_Dataset.html}} 

Before applying our method to the Endeavour dataset, we tuned our RANSAC-based ego-velocity estimator, synchronized the radar measurement timestamps and removed zero-velocity measurement pairs.
For RANSAC, the inlier and outlier thresholds were set to 40\% of the number of measured reflections and 0.025 m/s, respectively. 
These thresholds were determined using the radar and GNSS velocity data from \texttt{East\_2}.
To temporally synchronize the data streams, we aligned the radar measurement timestamps using linear interpolation.
Finally, we removed ego-velocity measurement pairs with magnitudes less than 0.05 m/s to improve the signal-to-noise ratio in the calibration problem.

Estimating the transforms for Olutomilayo et al.\ \cite{2021_Olutomilayo_Extrinsic} required two pre-processing steps. 
First, we identified stationary radar measurements using the RTK GNSS data and removed points that were observed less than five times.
Next, we expressed (using the Endeavour parameters) the radar point clouds in a common reference frame and associated points that were within a 10 cm threshold. 
Finally, the extrinsic transforms from Olutomilayo et al.\ \cite{2021_Olutomilayo_Extrinsic} and Endeavour were chained together to compute the rotation angles and translation axes relative to \texttt{Near\_5}.

\begin{table}[t!]
\scriptsize
\centering
\caption{$\theta_t$ and $\theta_{ba}$ parameters estimated by each method on \texttt{East2}. Radar $a$ is \texttt{Near\_5} for every parameter given.}
\begin{threeparttable}
	\begin{tabular}{lc@{\hspace{0.8\tabcolsep}}cc@{\hspace{0.8\tabcolsep}}cc@{\hspace{0.8\tabcolsep}}cc@{\hspace{0.8\tabcolsep}}c}
		\toprule
		& \multicolumn{2}{c}{\texttt{Near\_1}} & \multicolumn{2}{c}{\texttt{Near\_2}} & \multicolumn{2}{c}{\texttt{Near\_3}} & \multicolumn{2}{c}{\texttt{Near\_4}}\\
		\cmidrule(lr){2-3} \cmidrule(lr){4-5} \cmidrule(lr){6-7} \cmidrule(lr){8-9}
		Method & $\theta_t$ & $\theta_{ba}$ & $\theta_t$ & $\theta_{ba}$ & $\theta_t$ & $\theta_{ba}$ & $\theta_t$ & $\theta_{ba}$	\\
		\midrule
		Endeavour & 1.71 & -1.57 & 1.44 & 1.59 & 2.73 & -1.58 & 0.33 & 1.56 \\
		Olutomilayo \cite{2021_Olutomilayo_Extrinsic} &  1.36 & -1.60 & 1.41 & 1.58 & 2.16 & -1.61 & 0.39 & 1.56 \\
		Ours & 1.74 & -1.58 & 1.41 & 1.61 & 2.79 & -1.58 & 0.35 & 1.56 \\
		\bottomrule
	\end{tabular}
	\begin{tablenotes}
		\item *All angles are in radians.
	\end{tablenotes}
\end{threeparttable}
\label{tab:params}
\vspace{-4mm}
\end{table}

\Cref{tab:params} shows that our parameters are within 3$^\circ$ of the provided parameters while the parameters from Olutomilayo et al.\ \cite{2021_Olutomilayo_Extrinsic} deviate very significantly.
This deviation is due to the narrow overlap between the fields of view of some of the radar pairs, which results in sparse overlapping point clouds.
Often, this systematic issue results in the data collection runs having insufficient information for the method in \cite{2021_Olutomilayo_Extrinsic} to operate properly (see \Cref{tab:success-rate}). 

\begin{table}[b!]
\vspace{-3mm}
\scriptsize
\centering
\caption{Identifiability of Olutomilayo et al.\ \cite{2021_Olutomilayo_Extrinsic} for each run and radar pair in the Endeavour dataset. Our algorithm is identifiable for each run and radar pair.}
\begin{threeparttable}
    \begin{tabular}{lc@{\hspace{0.9\tabcolsep}}c@{\hspace{0.9\tabcolsep}}c@{\hspace{0.9\tabcolsep}}c@{\hspace{0.9\tabcolsep}}c@{\hspace{0.9\tabcolsep}}c}
        \toprule
        & \multicolumn{6}{c}{Data Collection Run}\\
        \cmidrule(lr){2-7}
        Radar Pairs & \texttt{East1} & \texttt{East2} & \texttt{Mid1} &\texttt{Mid2} & \texttt{West1} & \texttt{West2}\\
        \midrule
        \texttt{Near\_1}--\texttt{Near\_3} & \checkmark & \checkmark & \checkmark & \checkmark & \checkmark & \checkmark\\
        \texttt{Near\_2}--\texttt{Near\_4} & \checkmark & \checkmark & \checkmark & \checkmark & \checkmark & --- \\
        \texttt{Near\_3}--\texttt{Near\_5} & --- & \checkmark$^\dag$ & --- & --- & --- & --- \\
        \texttt{Near\_4}--\texttt{Near\_5} & --- & \checkmark & --- & --- & --- & --- \\
        \bottomrule
    \end{tabular}
    \begin{tablenotes}
    	\item $\dag$ This problem is only identifiable using features that appear in less than 5\% of measurements.
    \end{tablenotes}
\end{threeparttable}
\label{tab:success-rate}
\end{table}

We use the mean velocity error magnitude to evaluate the accuracy of the calibration parameters given in \Cref{tab:params}.
The velocity error of a radar measurement pair, $\Vector{h}_{r,a}^j$ and $\Vector{h}_{r,b}^j$, is $\Vector{e}_{r,b}^j$ from \Cref{eqn:errors}, where  $\Vector{v}_a^j = \Vector{h}_{r,a}^j$, $\theta_{ba}$ is the estimated rotation, $\theta_t$ is the estimated translation axis, and $\omega_\gamma^j$ is the value that minimizes the magnitude of $\Vector{e}_{r,b}^j$.
\Cref{tab:results} shows the mean velocity error magnitude for each run, radar pair, and set of parameters.
Our parameters yield lower velocity errors in almost all cases, reducing the mean  error magnitude for the \texttt{Near\_5}--\texttt{Near\_4} radar pair by over 1 cm/s.
\begin{table*}[t]
\scriptsize
\centering
\caption{Endeavour mean velocity error magnitude for each run excluding \texttt{East2}, which was used to estimate the calibration parameters. The errors presented below show how consistent the estimated parameters are when explaining the velocity vector field of a system with a unit moment arm.}
\begin{threeparttable}
	\begin{tabular}{lc@{\hspace{0.8\tabcolsep}}c@{\hspace{0.8\tabcolsep}}cc@{\hspace{0.8\tabcolsep}}c@{\hspace{0.8\tabcolsep}}cc@{\hspace{0.8\tabcolsep}}c@{\hspace{0.8\tabcolsep}}cc@{\hspace{0.8\tabcolsep}}c@{\hspace{0.8\tabcolsep}}c}
		\toprule
		& \multicolumn{3}{c}{\texttt{Near\_5}--\texttt{Near\_1}} & \multicolumn{3}{c}{\texttt{Near\_5}--\texttt{Near\_2}} & \multicolumn{3}{c}{\texttt{Near\_5}--\texttt{Near\_3}} & \multicolumn{3}{c}{\texttt{Near\_5}--\texttt{Near\_4}}\\
		\cmidrule(lr){2-4} \cmidrule(lr){5-7} \cmidrule(lr){8-10} \cmidrule(lr){11-13}
		Data & Endeavour & Olutomilayo \cite{2021_Olutomilayo_Extrinsic} & Ours &	Endeavour & Olutomilayo \cite{2021_Olutomilayo_Extrinsic} & Ours & Endeavour & Olutomilayo \cite{2021_Olutomilayo_Extrinsic} & Ours & Endeavour & Olutomilayo \cite{2021_Olutomilayo_Extrinsic} & Ours\\
		\midrule
		\texttt{Mid1} & 0.0218 & 0.0591 & \textbf{0.0175} & 0.0232 & 0.0242 & \textbf{0.0173} & 0.0185 & 0.0928 &  \textbf{0.0173} & 0.0411 & 0.0334 & \textbf{0.0184} \\
		\texttt{Mid2} & 0.0215 & 0.0580 & \textbf{0.0170} & 0.0228 & 0.0235 & \textbf{0.0166} & \textbf{0.0195} & 0.0885 & 0.0288 & 0.0289 & 0.0240 & \textbf{0.0140} \\
		\texttt{East1} & 0.0206 & 0.0657 & \textbf{0.0152} & 0.0208 & 0.0212 & \textbf{0.0157} & 0.0152 & 0.0846 & \textbf{0.0139} & 0.0305 & 0.0249 & \textbf{0.0121}\\
		\texttt{West1} & 0.0206 & 0.0574 & \textbf{0.0175} & 0.0247 & 0.0249 & \textbf{0.0179} & 0.0151 & 0.1323 & \textbf{0.0146} & 0.0354 & 0.0285 & \textbf{0.0149}\\
		\texttt{West2} & 0.0210 & 0.0558 & \textbf{0.0176} & 0.0259 & 0.0266 & \textbf{0.0192} & 0.0189 & 0.1320 & \textbf{0.0178} & 0.0515 & 0.0411 & \textbf{0.0224}\\
		\bottomrule
	\end{tabular}
	\begin{tablenotes}
		\item *All values are in m/s.
	\end{tablenotes}
\end{threeparttable}
\label{tab:results}
\vspace{-2mm}
\end{table*}

Due to the configuration of radar pairs \texttt{Near\_3}--\texttt{Near\_1} and \texttt{Near\_4}--\texttt{Near\_2}, the ego-motion of the vehicle driving forward aligns with the translation axes of these pairs, which, in theory, should make the calibration data poorly conditioned.
However, these radar pairs are mounted on the periphery of the vehicle, so any angular velocity induces unaligned ego-motion measurements.
Carrying out calibration on data from pairs \texttt{Near\_3}--\texttt{Near\_1} and \texttt{Near\_4}--\texttt{Near\_2}, with initial calibration parameters greater than 20$^\circ$ from the Endeavour values, results in estimated parameters that are consistently within 3$^\circ$ of the Endeavour values; this indicates that the problem is not poorly conditioned.

By including a third sensor that is able to measure angular velocity, we can estimate the (metric) scale of the translation between the radars, without requiring the exact extrinsic transform of the third sensor to be known.
The magnitude of the angular velocity of a rigid body is the same for all points on the body, allowing us to match the unscaled radar estimate to the angular velocity of the third sensor.
For example, assuming that the z-axis of an on-board GNSS receiver is roughly perpendicular to the sensing plane of the radar units, we can apply constant-acceleration smoothing and linear interpolation of the GNSS pose measurements to estimate angular velocity.
We tried this approach on the Endeavour dataset.
After removing measurement pairs with angular velocity magnitudes less than 0.1 rad/s, we computed the translation estimates.
\Cref{tab:trans-results} shows that, in most cases, the metric translation values recovered by our algorithm are closer to the ground-truth Endeavour dataset values than those estimated by Olutomilayo et al.\ \cite{2021_Olutomilayo_Extrinsic}.
While the sign of the translation may still be positive or negative (i.e., one z-axis could be inverted), this information can be easily determined from a rough model of the system. 

\begin{table}[b!]
\vspace{-2mm}
\scriptsize
\centering
\caption{Estimated translation magnitude for each method on \texttt{East2}. Radar $a$ is \texttt{Near\_5} for every parameter given. The bolded values are the translation magnitudes closest to the Endeavour parameters.}
\begin{threeparttable}
\begin{tabular}{lcccc}
	\toprule
	Method & \texttt{Near\_1} & \texttt{Near\_2} & \texttt{Near\_3} & \texttt{Near\_4} \\
	\midrule
	Endeavour & 5.68 & 5.82 & 0.83 & 0.86\\
	Olutomilayo \cite{2021_Olutomilayo_Extrinsic} & 4.87 & \textbf{6.04} & 0.47 & 0.96 \\
	Ours & \textbf{6.08} & 5.54 & \textbf{0.77} & \textbf{0.95} \\
	\bottomrule
\end{tabular}
\begin{tablenotes}
	\item *All values are in m.
\end{tablenotes}
\end{threeparttable}
\label{tab:trans-results}
\end{table}

\section{Conclusion}
\label{sec:conclusion}

In this paper, we presented a 2D radar-to-radar extrinsic calibration algorithm that uses radar ego-velocity data only.
We proved that the yaw angle and the axis of translation between the sensors can be identified given sufficient excitation.
Using simulations, we demonstrated that our calibration method is robust to varying levels of radar measurement noise and that we are able to improve the raw radar ego-velocity estimates.
Finally, we showed, using data from a vehicle, that our algorithm was more reliable and accurate than a state-of-the-art method.

There are multiple potential directions for future research. 
Our approach could be extended to pairs of 3D radar sensors, similar to those discussed in Wise et al.\ \cite{wise_continuous-time_2021}.
Another possibility is to perform temporal calibration using ego-velocity estimates, which could simplify the estimation problem for some systems.

\bibliographystyle{IEEEtran}
\bibliography{refs}

\end{document}